\documentclass[letterpaper, 10 pt, conference]{ieeeconf}  % Comment this line out if you need a4paper

\IEEEoverridecommandlockouts                              % This command is only needed if 
\usepackage{amsmath} % assumes amsmath package installed
\usepackage{amssymb}  % assumes amsmath package installed

\usepackage{subfig}
\usepackage{graphicx}
\usepackage[colorlinks = true, urlcolor  = blue]{hyperref}
\usepackage{cite}
\usepackage{booktabs}
\usepackage{dsfont}
\usepackage{multirow}

\usepackage{xcolor}
\usepackage{soul}

\newcommand\blfootnote[1]{%
  \begingroup
  \renewcommand\thefootnote{}\footnote{#1}%
  \addtocounter{footnote}{-1}%
  \endgroup
}

\title{\LARGE \bf
DexTouch: Learning to Seek and Manipulate Objects \\ with Tactile Dexterity  
}

\author{
Kang-Won Lee$^{1}$, 
Yuzhe Qin$^{2}$,
Xiaolong Wang$^{2}$,
and Soo-Chul Lim$^{1}$
\thanks{This research was supported by the Technology Innovation Program (Grant No. 20016252) funded by the Ministry of Trade, Industry and Energy (MOTIE, Korea) and the MOTIE, under the Fostering Global Talents for Innovative Growth Program (P0017307) supervised by the Korea Institute for Advancement of Technology (KIAT)}% <-this % stops a space
\thanks{$^{1}$the Department of Mechanical, Robotics and Energy Engineering, Dongguk University, Seoul 04620, South Korea (e-mail : leekw@dgu.ac.kr, limsc@dongguk.edu)}%
\thanks{$^{2}$University of California San Diego, CA, USA}%
}

\begin{document}
% \maketitle

\twocolumn[{%
\renewcommand\twocolumn[1][]{#1}%
\maketitle
\begin{center}
    \vspace{-7mm}
    \centering
    \captionsetup{type=figure}
    \includegraphics[width=0.8\linewidth]{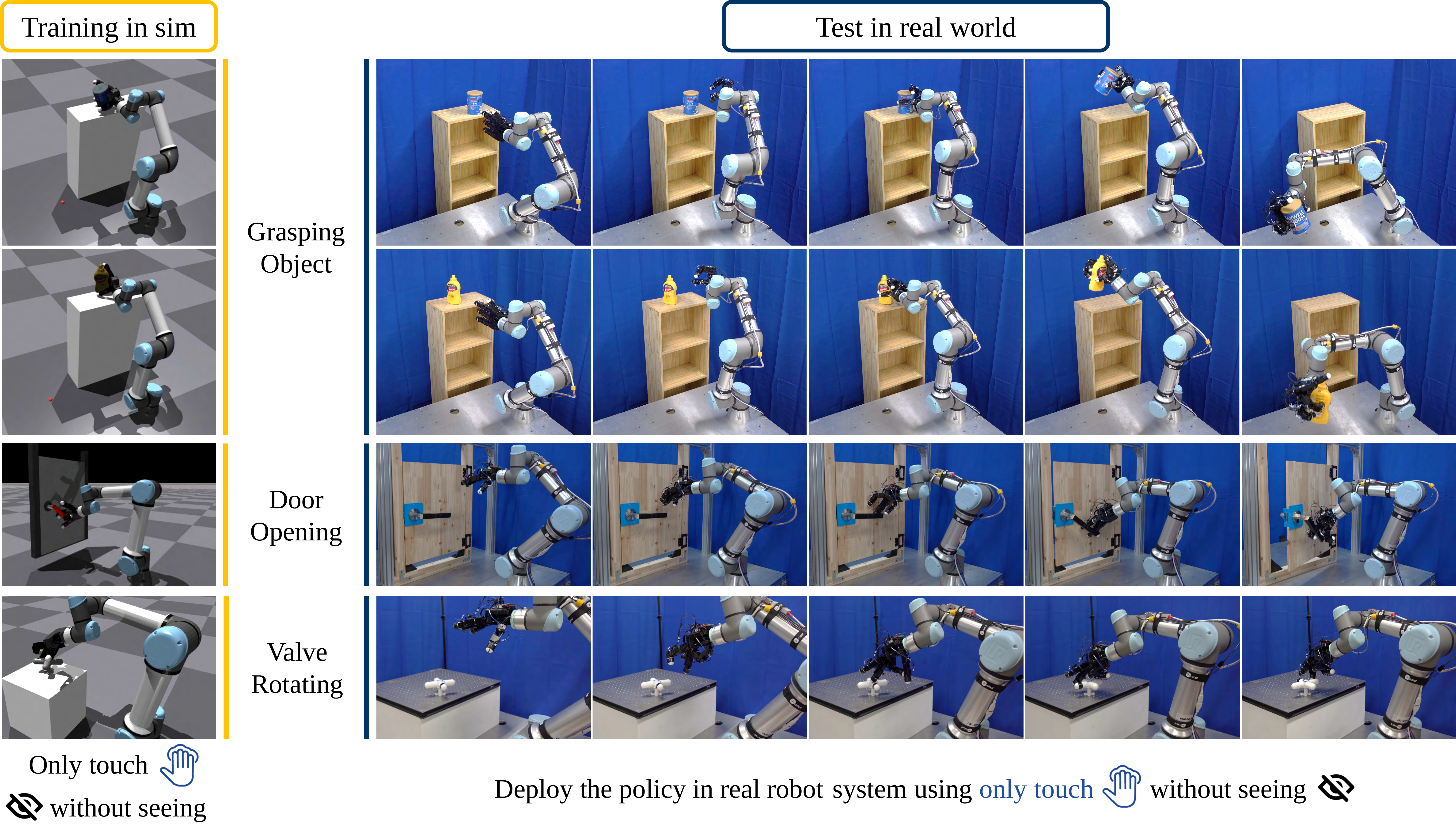}
    \vspace*{-2mm}
    \captionof{figure}{
    We propose \textbf{DexTouch}, a dexterous robotic system with tactile-based blind manipulation. The robotic system consisted of a UR5e arm and an AllegroHand with 16 attached touch sensors. Policies were trained in simulation and then deployed to the real environment without fine-tuning.
    }
    \label{fig:teaser}
\end{center}
}]

\blfootnote{
This article was published in IEEE Robotics and Automation Letters. Digital
Object Identifier (DOI): 10.1109/LRA.2024.3478571 \\
This research was supported by the Technology Innovation Program (Grant No. 20016252) funded by the Ministry of Trade, Industry and Energy (MOTIE, Korea) and the MOTIE, under the Fostering Global Talents for Innovative Growth Program (P0017307) supervised by the Korea Institute for Advancement of Technology (KIAT) \textit{(Corresponding author : Xiaolog Wang, Soo-Chul Lim)}\\
\textsuperscript{$1$} The authors are the Dongguk University, Seoul 04620, South Korea (e-mail: leekw@dgu.ac.kr, limsc@dongguk.edu) \\
\textsuperscript{$2$} The authors are the University of California San Diego, La Jolla, CA 92093, USA (e-mail: y1qin@eng.ucsd.edu, xiw012@ucsd.edu) \\
© 2024 IEEE. Personal use of this material is permitted. Permission from
IEEE must be obtained for all other uses, in any current or future media,
including reprinting/republishing this material for advertising or promotional
purposes, creating new collective works, for resale or redistribution to servers
or lists, or reuse of any copyrighted component of this work in other works.
}

\thispagestyle{empty}
\pagestyle{empty}

%%%%%%%%%%%%%%%%%%%%%%%%%%%%%%%%%%%%%%%%%%%%%%%%%%%%%%%%%%%%%%%%%%%%%%%%%%%%%%%%
\vspace{-3mm}
\begin{abstract}

The sense of touch is an essential ability for skillfully performing a variety of tasks, providing the capacity to search and manipulate objects without relying on visual information.
In this paper, we introduce a multi-finger robot system designed to manipulate objects using the sense of touch, without relying on vision.
For tasks that mimic daily life, the robot uses its sense of touch to manipulate randomly placed objects in dark.
The objective of this study is to enable robots to perform blind manipulation by using tactile sensation to compensate for the information gap caused by the absence of vision, given the presence of prior information.
Training the policy through reinforcement learning in simulation and transferring the trained policy to the real environment, we demonstrate that blind manipulation can be applied to robots without vision.
In addition, the experiments showcase the importance of tactile sensing in the blind manipulation tasks.
Our project page is available at \href{https://lee-kangwon.github.io/dextouch/}{https://lee-kangwon.github.io/dextouch/}

\end{abstract}

%%%%%%%%%%%%%%%%%%%%%%%%%%%%%%%%%%%%%%%%%%%%%%%%%%%%%%%%%%%%%%%%%%%%%%%%%%%%%%%%
% \vspace*{0.1mm}
\section{Introduction}
Consider a situation of retrieving something from inside of a cabinet or from a tall cupboard where shelves are not visually observable. Despite the fact that we cannot see the object in the high cupboard and don't know its exact position, we can easily seek and retrieve them. Likewise, we can seek a switch in a dark room or open the door to leave. The reason humans can perform these tasks in daily life is that we can perceive our surroundings or interacting objects through the sense of touch, even when visual information is unavailable. 

In many previous studies, touch has primarily been used for dexterous manipulation in robots\mbox{~\cite{sobinov2021neural, dex_overview, andrychowicz2020learning, chavan2020sampling, li2020review}}. However, the use of the tactile sense is not limited to improving performance in manipulation tasks. Robots primarily rely on vision for information while performing manipulation tasks. However, if vision is limited, accurate information cannot be known in addition to the information known in advance, which may cause difficulties in manipulation.
As in the previous example, when the approximate location is known but the exact location cannot be visually determined due to limitations in available vision, an information gap occurs. The tactile sense can help compensate for gaps in these situations, aiding the robot in performing tasks successfully. 

Our purpose is to enable blind manipulation for robots by compensating for information gaps using tactile information. Previous studies have demonstrated that the tactile sense is useful for blind grasping, which involves grasping objects in an environment without vision. However, vision can be limited in many situations, so it is necessary to extend it to diverse manipulation tasks. In addition, a multi-finger robot hand was utilized to perform these various tasks.

We present \textbf{DexTouch}, a tactile-based multi-finger robot system capable of manipulating objects in environments where vision is not available.
We utilize binary tactile state information to overcome the Sim2Real gap, following previous research~\cite{touch-dexterity}.
The tactile sensors are attached to one side of the hand, covering the fingertips, links, and palm, and equipped with 16 the Force Sensing Resistor (FSR) sensors, as illustrated in Fig.~\ref{fig:teaser}. 
With this system setup, we focus on solving the task of manipulating objects in an environment without vision. To do this, we set three types of tasks that are relevant to our daily lives. 
These tasks are generally performed using vision, however when the position of the object cannot be accurately determined due to visual limitations, the task becomes significantly more difficult.

We use Reinforcement Learning (RL) on the IsaacGym simulator{~\cite{makoviychuk2021isaac}} to train policies for complex movements of a robot hand and arm, which can be deployed in real environments. 
We demonstrate that RL can be used to manipulate objects in environments where visual observations are not possible by leveraging observable tactile sensations.
In our experiments, we test a real world-system for three types of the tasks.
The results demonstrate that the touch-based approach is useful for blind manipulation and can be effectively applied in real-world.
Additionally, we conduct ablations on our system to validate our design, including tasks such as disabling the touch sensor and adjusting the sensitivity of the binary force sensor.

\section{Related Work}
\textbf{Dexterous Manipulation}
allows robots to perform a variety of tasks, allowing them to be applied to a wide range of fields~\cite{dex_overview, andrychowicz2020learning, chavan2020sampling}. Particularly, the implementation of the dexterous manipulative ability of robots that manipulate objects is receiving a lot of attention~\cite{andrychowicz2020learning, Bhatt-RSS-21, morgan2022complex}. To address the challenge of dexterous manipulation, model-based approaches to modeling robots and interaction systems were initially proposed~\cite{kumar2014real, bai2014dexterous}. However, approaches utilizing classical controls are based on designed dynamical models, which limits their usefulness for scaling to more complex tasks. 
To overcome these limitations, methods using deep reinforcement learning have recently been receiving attention~\cite{handa2023dextreme, chen2022system, qin2023dexpoint, qin2022dexmv, chen2023visual}. Effective learning and high task success rates can be achieved by utilizing large-scale parallel learning using simulation. 
Most methods rely on visual input and address issues like occlusion, focusing primarily on dexterous in-hand manipulation limited to small movements. In contrast, our system uses touch sensors to learn movements of robotic arms and hands, aiming to expand into tasks that mimic daily life.

\textbf{Tactile-based Manipulation}
can compensate for the loss of information due to the absence of vision\mbox{~\cite{chebotar2014learning, ko2023vision, lee2021toward}} and enable manipulation tasks that consider the properties of objects difficult to grasp with vision alone{~\cite{liu2017recent}}. 
% \hl{In particular, tactile detection has been useful for stably grasping large or unfamiliar objects{~\cite{mittendorfer2013general}}. In addition, tactile sense can be utilized to effectively grasp objects in motion{~\cite{lynch2021adaptive}} and enable stable gripping of objects without vision, known as blind grasping\mbox{~\cite{dang2011blind, shaw2018tactile}}. 
% These methods model the system or construct control loops with tactile feedback to ensure stable grip on objects. However, due to constraints associated with modeling, their extension to complex tasks is limited. In contrast, we propose a RL-based approach that enables complex object manipulation.}
In particular, tactile detection has been useful for stably grasping large or unfamiliar objects{~\cite{mittendorfer2013general}}. In addition, tactile sense can be utilized to effectively grasp objects in motion{~\cite{lynch2021adaptive}} and enable stable gripping of objects without vision, known as blind grasping\mbox{~\cite{dang2011blind, shaw2018tactile}}. 
These methods model the system or construct control loops with tactile feedback to ensure stable grip on objects. However, due to constraints associated with modeling, their extension to complex tasks is limited. In contrast, we propose a RL-based approach that enables complex object manipulation.

Various approaches have been proposed to identify the surrounding environment and objects in the absence of vision. In particular, a method for identifying unknown objects through iterative grasping has also been studied\mbox{~\cite{dragiev2013uncertainty, pai2023tactofind}}. An approach that models the environment by utilizing continuous contact between a robot and an object has also been proposed {~\cite{schneider2022active}}. In addition, research has been conducted on recognizing transparent objects, which are difficult to explore visually, using tactile sense{~\cite{murali2023touch}}, and on estimating the accurate 6-DOF pose of an object in a cluttered environment by utilizing both vision and tactile sense{~\cite{murali2022active}}. However, these tactile-based recognition methods have primarily been implemented using grippers with low degrees of freedom. In contrast, we propose a method for performing blind manipulation by applying tactile sensing to a multi-finger robot hand.

\section{Tactile Robotic System}
\begin{figure*}
    \centering
    \includegraphics[width=0.9\linewidth]{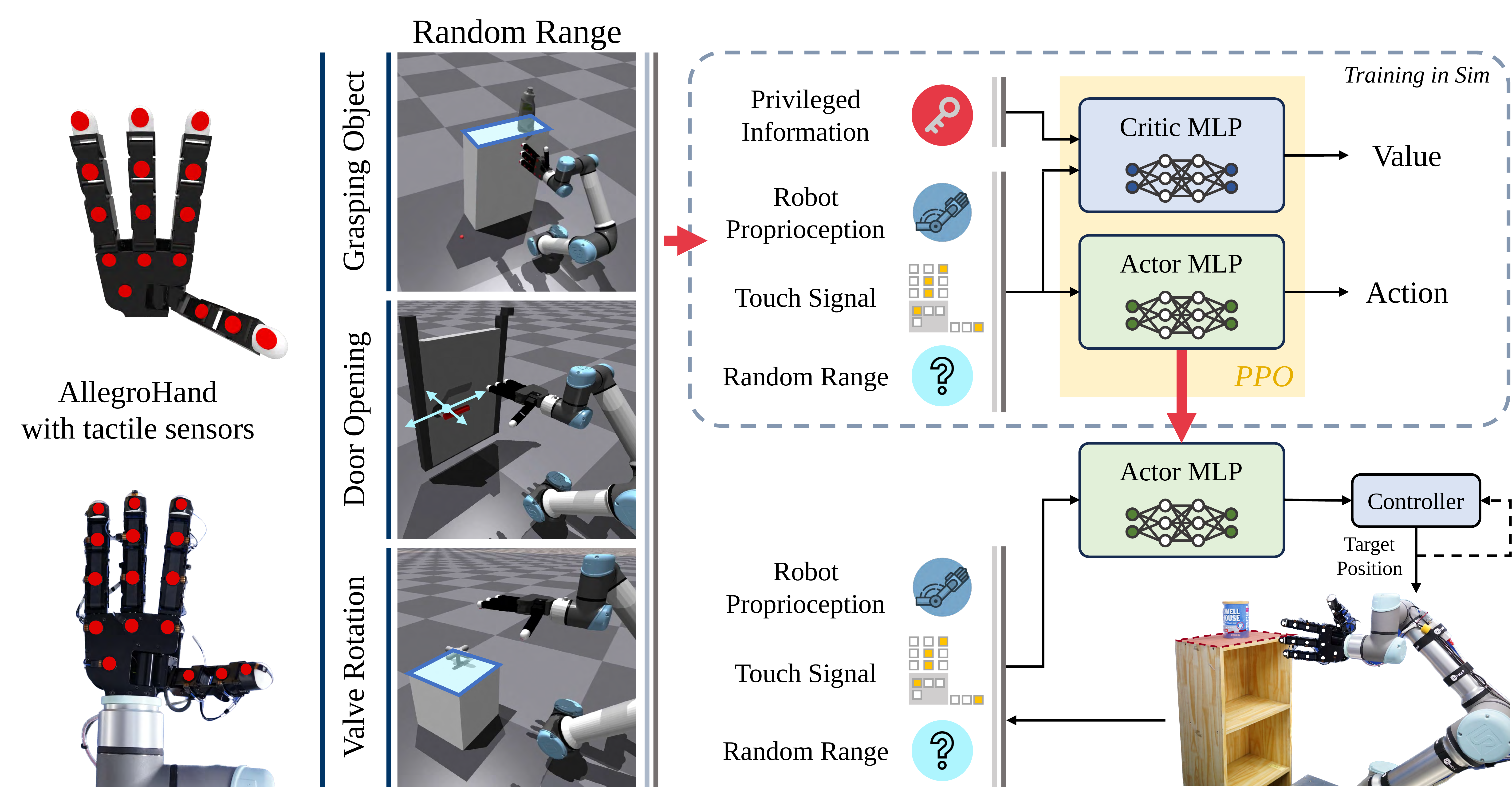}
    \caption{Pipeline of the system. Left: visualization of the Allegro hand with tactile sensors (red indicators) in simulation and real-world. Right: visualization for all three tasks in simulation and training pipeline. The blue areas and arrows represent the random range of the target object.
    The state contains robot proprioception, tactile information, and task information like goal position and random range from origin state. Then the policy uses this state to get actions.}
    \label{fig:pipeline}
    \vspace{-5mm}
\end{figure*}

\subsection{System Setup}
Our system is a multi degree-of-freedom (DoF) robot with a total of 22-DoF, consisting of a 6-DoF UR-5e and a 16-DoF AllegroHand. The robot hand has a total of 16 FSR sensors, with 3 sensors attached to each finger and 4 sensors on the palm, as illustrated by the white dots in Fig.~\ref{fig:pipeline} and as suggested by~\cite{touch-dexterity}. 
Using an STM32F103 microcontroller, the sensor's voltage signal is measured at a sampling rate of 125 Hz, passed through a low-pass filter to remove noise, and then transmitted to the main controller.
Binary signals are used to reduce the gap between simulation and real robots and simplify the Sim2Real transfer procedure. For this purpose, preprocessing is performed to convert the transmitted voltage signal into a binary contact signal according to a selected threshold $\theta_{th}$ before being used. %as input to the control.

IssacGym simulator~\cite{makoviychuk2021isaac} was used for the training of our tactile based robot system. The simulation settings for each task are visualized in Fig.~\ref{fig:teaser} (left). A multi-DoF robotic arm-hand system was implemented in the simulation, identical to a real robot. Sixteen virtual contact sensors are configured in the simulation and $\|F\|$ calculated from the net contact force $F = [F_x, F_y, F_z]$ measured for each step is used as the contact force. Then, using a selected threshold $\Tilde{\theta}_{th} = 0.01$ $\mathrm{N}$ the binary contact signal is calculated in the same way as in real-world environments. The control frequency was set to 10 Hz in both real-world environments and simulations.

\subsection{Task Description}
\begin{figure}
    \centering
    \includegraphics[width=0.85\linewidth]{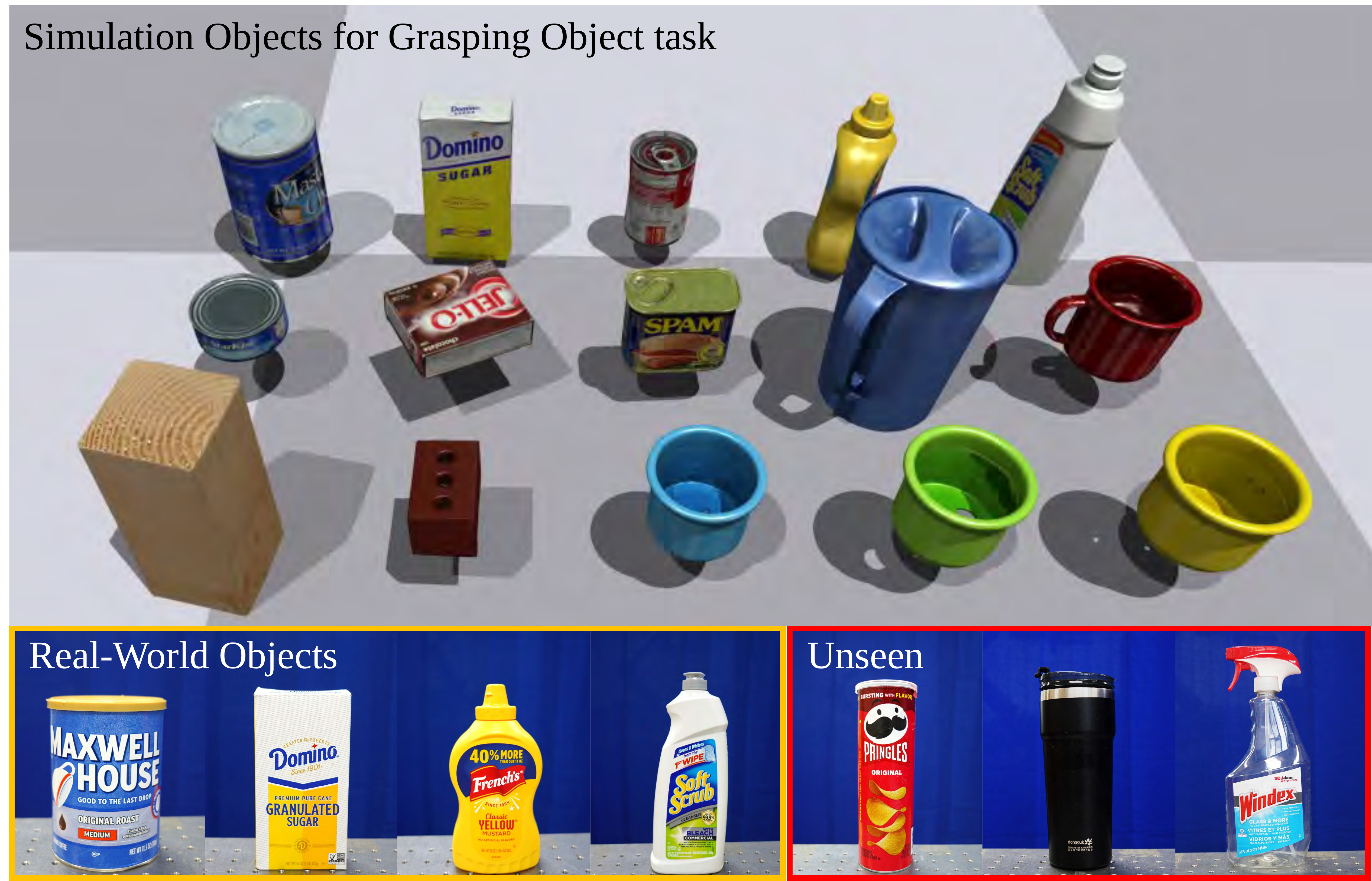}
    \caption{The object sets used in grasping object task. The objects in red box are the unseen objects.}
    \label{fig:objects}
    \vspace{-7mm}
\end{figure}

We study dexterity using a robot's sense of touch through various types of object manipulation tasks aimed at manipulating objects without vision.
In this paper, we mainly focus on three distinct types of benchmark tasks that are relevant to our daily lives as illustrated in Fig.~\ref{fig:pipeline}.
\textbf{(i) Grasping Object.}
In this task, inspired by human behavior of retrieving items from a high, invisible shelf, the robot is required to utilize its sense of touch to grasp objects randomly positioned on the table and then bring them to the target point without dropping them. Furthermore, the policy should demonstrate that the robot can also manipulate multiple objects with different shapes.
Fifteen objects were selected from the YCB dataset~\cite{calli2015ycb}, as depicted in the Fig.~\ref{fig:objects}, and experiments were conducted on seven of them in a real-world environment. Three of these are unseen objects.
\textbf{(ii) Door Opening.}
In this task, the robot needs to locate a door handle that is positioned randomly about the x and y-axes, rotate the door handle, and pull it to open the door. In an invisible environment, the robot should use its sense of touch to locate door handles and learn complex movements to open the door.
\textbf{(iii) Valve Rotating.}
The purpose of this task is to locate the position of a valve located randomly on a plane and rotate the valve clockwise. Compared to objects in other tasks, the size of the valve is small, requiring more sophisticated exploration ability of the robot.

\section{Learning Tactile Manipulation}

\subsection{Problem Formulation}
Our tasks are formulated as a Markov Decision Process (MDP) $\mathcal{M} = (\mathcal{S}, \mathcal{A}, \mathcal{R}, \mathcal{P})$. Here, $\mathcal{S}$ is the state space, $\mathcal{A}$ is the action space, $\mathcal{R}$ is the reward function, and $\mathcal{P}$ is the transition dynamics. 
The objective of robot agent is to find optimal policy $\pi$ to maximize the $\gamma$ discounted return $\sum_{t=0}^T \gamma^t r_t$.
The agent observes state $s_t$ at each time step $t$. Then, it take action $a_t = \pi(s_t)$ calculated by policy $\pi$ and receive a reward $r_t = \mathcal{R}(s_t, a_t, s_{t+1})$. During this process, the agent doesn't know $\mathcal{R}$ and $\mathcal{P}$. An episode terminates when the agent exceeds the maximum number of steps $T$ or achieves the goal of task or reset conditions are achieved.
\subsubsection{State}
The state of the system consists of the joint position $q_t \in \mathbb{R}^{22}$, joint velocity $\dot{q}_t \in \mathbb{R}^{22}$ of 6-DoF UR-5e arm and 16-DoF Allegro hand, the binary contact signal $o_t \in \{0, 1\}^{16}$, 7-Dof palm pose, 3-Dof linear and 3-Dof angular velocity $p_{plam} \in \mathbb{R}^{13}$, 4-fingertip positions relative to robot palm $p_{tips} \in \mathbb{R}^{12}$, and task information $I$ for each task. 
$I$ consists of the goal position $(goal_{x}, goal_{y}, goal_{z})$ and the length of random range $(range_{x}, range_{y})$ $I \in \mathbb{R}^{5}$ in \textit{Grasping Object}. In case of \textit{Door Opening} and \textit{Valve Rotating}, I consists of only the length of random range $(range_{x}, range_{y})$ $I \in \mathbb{R}^{2}$.
The length of random range refers to the size of the area where objects are randomly spawned. The x-axis corresponds to the horizontal direction of the robot hand, while the y-axis corresponds to the vertical direction. This prior information is the minimum necessary for the robot to recognize that an object exists. Based on this, the robot learns how to manipulate the object using touch.

\subsubsection{Action}
At each step, the action calculated by the policy network is a normalized vector $a_t \in \mathbb{R}^{22}$ that consists of two parts, 6-DoF for robot arm and 16-DoF for robot hand. For the robot arm, the first 6-D vector was utilized as a position control command for each joint of the robot arm. In the case of the robot hand, a joint position controller is employed to command the position target of its 16 finger joint angles. Then, both the hand and arm are controlled by PD controllers.

\begin{figure}
    \centering
    \includegraphics[width=\linewidth]{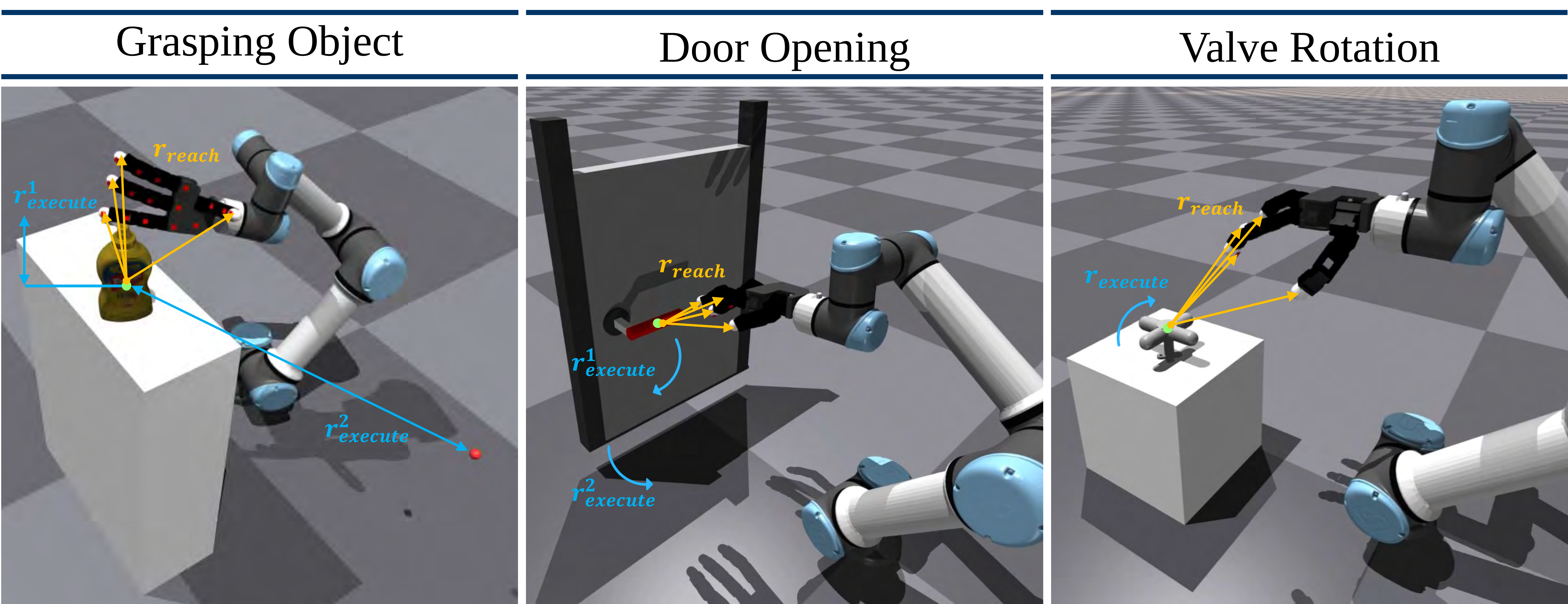}
    \caption{Visualization of the reward for each task. Yellow arrows represent reward $r_{reach}$ based on the distance between each fingertip and the target object. Blue arrows represent reward $r_{execute}$ based on manipulative actions to perform each task. The total reward function is the sum of the two rewards.}
    \label{fig:reward}
    \vspace{-6mm}
\end{figure}

\subsection{Reward Design}
For a successful application of reinforcement learning, the density of rewards should be enough to facilitate exploration by the agent. At the same time, the agent should be focused on achieving the goal without being distracted from taking actions that deviate from the final goal.
To this end, a reward function is proposed that decomposes the task into two phases: reaching the target object and executing the purposeful manipulation task, as illustrated in Fig.~\ref{fig:reward}. Additionally, inspired by~\cite{petrenko2023dexpbt}, a reward function is designed such that the agent can only receive rewards for movements toward the goal.

\textbf{Reaching the target object.}
To encourage the robot to get closer to the target object, the reach reward $r_{reach}$ was designed as follows:
\begin{equation}
r_{reach}=\underset{finger}{\sum} \alpha_{reach}\times\max(d_{closest}-d, 0)
\label{equ:reach}
\end{equation}
This reward is common to all three tasks. Here, $d$ and $d_{closest}$ are both the distance between each fingertip and the target object. The target object is the object on the table in \textit{Grasping Object}, the door handle in \textit{Door Opening}, the center position of valve in \textit{Valve Rotating}. $d$ is the current distance, and $d_{closest}$ is the closest distance achieved during actions so far. At the beginning of the episode, set $d_{closest} = d$. $\alpha_{reach}$ is a relative reward weight.

\textbf{Executing the manipulation.}
Since the target manipulation varies for each task, a manipulation reward $r_{execute}$ was designed according to each task. First, $r_{execute}$ for \textit{Grasping Object} task is designed as follows:
\begin{multline}
     r_{execute}=(1-\mathds{1}_{picked})\times\alpha_{pick}\times h_{obj} + r_{picked} 
     \\ 
     + \mathds{1}_{picked} \times \alpha_{goal} \times \max(\Tilde{d}_{closest}-\Tilde{d}, 0)
\label{equ:grasping}
\end{multline}
In equation~\ref{equ:grasping}, the component $r_{execute}$ rewards the agent for picking up an object, lifting it off the table, and bringing it to the target point. Here, $\mathds{1}_{picked}$ is an indicator function that activates when the height of the object $h_{obj}$ relative to the table becomes 1 above the preset threshold of 10 $\mathrm{cm}$. At this moment the agent
receives an additional bonus reward $r_{picked}$. $\Tilde{d}$ and $\Tilde{d}_{closest}$ represent the distance between the object and the target point, where $\Tilde{d}$ is the current distance and $\Tilde{d}_{closest}$ is the closest distance achieved during attempt so far. $\alpha_{pick}$ and $\alpha_{goal}$ are relative reward weights.

For \textit{Door Opening} task, $r_{execute}$ is designed as follows:
\begin{align}
	r_{execute} & =(1-\mathds{1}_{rotated})\times\alpha_{rot}\times \max(\phi-\phi_{max}, 0) \nonumber\\
	    & +\mathds{1}_{rotated} \times \alpha_{open} \times \max(\psi-\psi_{max}, 0) \nonumber\\ 
	    & + r_{rotated} + r_{opened}
\label{equ:door}
\end{align}

In equation~\ref{equ:door}, the component $r_{execute}$ rewards the agent for actions such as rotating the door handle and opening the door. $\mathds{1}_{picked}$ is an indicator function that becomes 1 when the rotation angle of the door handle exceeds the preset threshold of 1.047 $\mathrm{rad}$ (60 $\mathrm{deg}$). $\phi$ and $\phi_{max}$ are the door handle angles, $\phi$ is the current door handle angle, and $\phi_{max}$ is the maximum door handle angle achieved during attempt. Similarly, $\psi$ and $\psi_{max}$ are the door angles, $\psi$ is the current door angle, and $\psi_{max}$ is the maximum door angle achieved. When the door handle is rotated beyond the threshold, reward $r_{rotated}$ is received, and when the door is opened beyond a preset threshold of 0.873 $\mathrm{rad}$ (50 $\mathrm{deg}$), reward $r_{opened}$ is received. $\alpha_{rot}$ and $\alpha_{open}$ are relative reward weights.

In case of \textit{Valve Rotating} task, $r_{execute}$ is designed as follows:
\begin{align}
	r_{execute} &= \alpha_{rot} \times \max(\theta-\theta_{max}, 0) + r_{success}
\label{equ:valve}
\end{align}

In equation~\ref{equ:valve}, the component $r_{execute}$ rewards the agent for the action of rotating the valve. It is similar to the reward for rotating the door handle in the previous task. When the valve is rotated more than 135 $\mathrm{deg}$, the agent receives a success bonus. $\alpha_{rot}$ is a relative reward weight.

Finally, to prevent unstable jerky movements of the robot, we add a simple adjustable penalty containing the L1 norm of the velocity of each joint to promote smooth movement of the robot agent.

\subsection{Training Procedure}
The policy is trained using experiences simulated in IsaacGym~\cite{makoviychuk2021isaac}, a highly parallelized GPU-accelerated physics engine. Both the policy and value networks consist of 3 multi-layer perceptrons (MLPs) with sizes 512, 256, and 128, respectively, and ELU~\cite{nair2010rectified} as the activation function. To train the control policy, the Proximal Policy Optimization (PPO)~\cite{schulman2017proximal} algorithm is employed with the following hyper-parameters: clipping $\epsilon=0.2$, discount factor $\gamma=0.99$ and 0.016 KL threshold. To reduce the difficulty of learning, asymmetric observation is applied to policy and value networks~\cite{handa2023dextreme}. Specifically, the value network includes privileged information such as the exact pose of the target object, linear and angular velocities, distance from the robot hand, and physical parameters, and includes information for calculating rewards such as indicator functions depending on the task.

For the IsaacGym simulation, 4096 parallel environments were used in $dt=0.01667$ $\mathrm{s}$ with 2 simulation sub-steps. The action generated by the policy network is executed over 6 steps, corresponding to a control frequency of 10Hz in real-world.

\section{Experiments}
In this part, our DexTouch system was compared to several baselines in both the simulation and real-environments. Experiments are conducted on the proposed tasks, which include grasping objects, opening doors, and rotating valves as defined in the method. Experiments and analyses are mainly conducted on the following perspectives: 

(i) For every task, the task success rate of the proposed system on randomly positioned target objects is evaluated. In particular, an ablate study is conducted on how tactile information affects the search of objects in an environment without vision.

(ii) For every task, transfer performance to the real world is evaluated. As in simulation, the success rate of each task is evaluated for objects randomly located within a random range. Details about the random area are shown in Table{~\ref{table:random_range}}. Objects were randomly positioned within the same area as in the simulator for all tasks.
For grasping objects, 30 trials were performed for each object. In door opening and valve rotation, 55 experiments were conducted with randomly positioned doors and valves.

\subsection{Baselines}
In experiments, our method is mainly compared with the following baselines.
\begin{itemize}
    \item [1)]
    \textbf{WO-Sensor (Without-Sensor).} This PPO policy is learned without any tactile information from the robot. 
    It should use only the robot’s proprioception to manipulate the target object during the task.
    \item [2)]
    \textbf{LQ-Sensor (Low Quality-Sensor).} The threshold of the tactile sensor to detect touch was set to 0.3 $\mathrm{N}$. In other words, by reducing the sensor's sensitivity, a lower-quality tactile sensor is employed to train the policy.
    \item [3)]
    \textbf{WO-PInfo (Without-Privileged Information).} This PPO policy is learned without privileged information such as the exact pose of the target object, linear and angular velocities, distance from the robot hand, and physical parameters, and includes information for calculating rewards such as indicator functions.
    \item [4)]
    \textbf{DA-Sensor (Deactivation-Sensor).} Training is conducted following the same procedure as the proposed method, with the only difference being the deactivation of the tactile sensor during evaluation. This policy is used for ablation purposes and to test the extent to which tactile information is used when applied in a real-world.
\end{itemize}

\begin{table}
\centering
\setlength{\abovecaptionskip}{0pt}
\setlength{\belowcaptionskip}{0pt}
\caption{
The object's initial position is uniformly randomly selected from the center of the area.}
\resizebox{\columnwidth}{!}{%
% \colorbox{yellow}{
\begin{tabular}{lccc}
\toprule[1.2pt]
Task                & $Range_{x}$ ($\mathrm{m}$)    & $Range_{y}$ ($\mathrm{m}$)    & Z-Pose (rad) \\ 
\midrule
Grasping Object     & $+\mathcal{U}(-0.30, 0.30)$   & $+\mathcal{U}(-0.15, 0.15)$   & $[-\pi, \pi]$ \\
Door Opening        & $+\mathcal{U}(-0.55, 0.55)$   & $+\mathcal{U}(-0.20, 0.20)$   & - \\
Valve Rotating      & $+\mathcal{U}(-0.30, 0.30)$   & $+\mathcal{U}(-0.30, 0.30)$   & $[-\pi, \pi]$ \\
\bottomrule[1.2pt]
\end{tabular}%
}
\label{table:random_range}
\vspace{-7mm}
\end{table}

In this paper, prior information about the environment is assumed. This information does not provide the exact location of the object, but only the range in which it is likely to exist. Since the height of the furniture or door generally does not change, randomness was applied to the vertical and horizontal distances from the robot, with the height remaining constant. In addition, this was set considering the operating range of the robot because the robot arm is fixed. This randomness was applied consistently in both the simulation and the real-world.

\subsection{Ablation of tactile sensors during policy training}
The Fig.~\ref{fig:result_plot} shows the success rate and episodic return of each task according to the training process of the proposed method and baselines and Table~\ref{table:results-sim} shows the learning results of each method. The findings are highlighted as follows:

\begin{figure*}[htbp]
    \centering
    \includegraphics[width=1\linewidth]{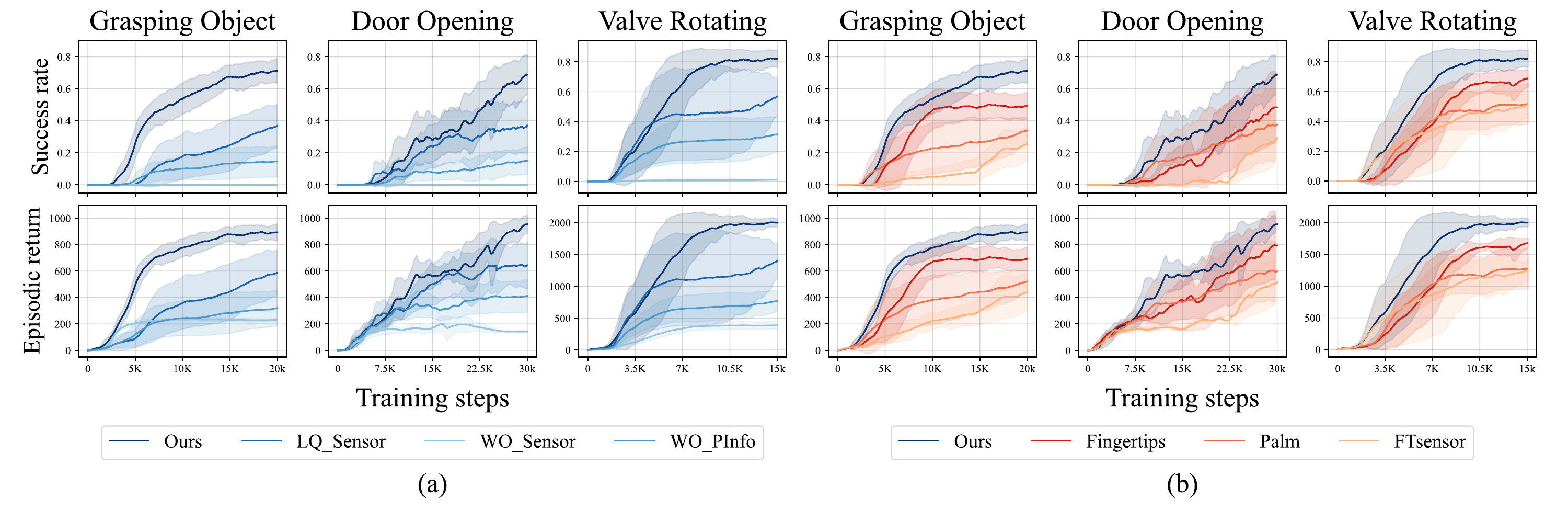}
    \vspace*{-6mm}
    \caption{Training process depending on (a) the attachment and sensitivity of the tactile sensor and (b) the location of the tactile sensor and other type of sensor.
    The results are averaged on 3 seeds, and the shaded area indicates the standard deviation.
    The x-axis is the training steps.
    The y-axis of the upper row is the success rate for each task and the lower row is the episodic return.
    }
    \label{fig:result_plot}
    \vspace{-2mm}
\end{figure*}

\begin{table*}[htbp]
\centering
\caption{
Performance on each manipulation task in simulation. The results are averaged over 3 policies trained on 3 seeds.}
\vspace{-1mm}
\resizebox{\textwidth}{!}{%
{\scriptsize
% \colorbox{yellow}{
\renewcommand{\arraystretch}{.95}
\begin{tabular}{lcccccc}
\toprule[1pt]
\renewcommand{\arraystretch}{2}
\multirow{2}{*}{Method} & \multicolumn{2}{c}{Grasping Object}                   & \multicolumn{2}{c}{Door Opening}                      & \multicolumn{2}{c}{Valve Rotation} \\
                        & Success rate              & Reward                    & Success rate              & Reward                    & Success rate              & Reward                    \\
\midrule
LQ-Sensor               & 0.37$_{\pm{0.14}}$        & 587.75$_{\pm{172.96}}$    & 0.37$_{\pm{0.17}}$        & 644.11$_{\pm{179.13}}$    & 0.58$_{\pm{0.13}}$        & 1400.99$_{\pm{284.58}}$   \\
WO-Sensor               & 0.0$_{\pm{0.0}}$          & 232.80$_{\pm{11.28}}$     & 0.0$_{\pm{0.0}}$          & 142.57$_{\pm{10.03}}$     & 0.02$_{\pm{0.01}}$        & 389.01$_{\pm{59.45}}$     \\
WO-PInfo                & 0.15$_{\pm{0.10}}$        & 320.37$_{\pm{134.71}}$    & 0.15$_{\pm{0.09}}$        & 411.89$_{\pm{123.49}}$    & 0.31$_{\pm{0.12}}$        & 771.19$_{\pm{168.35}}$    \\
\midrule
Fingertips              & 0.49$_{\pm{0.10}}$        & 693.12$_{\pm{89.73}}$     & 0.48$_{\pm{0.22}}$        & 792.18$_{\pm{258.47}}$    & 0.69$_{\pm{0.07}}$        & 1677.22$_{\pm{169.42}}$   \\
Palm                    & 0.36$_{\pm{0.09}}$        & 520.74$_{\pm{82.42}}$     & 0.38$_{\pm{0.23}}$        & 599.27$_{\pm{228.23}}$    & 0.53$_{\pm{0.14}}$        & 1273.58$_{\pm{320.24}}$   \\
F/Tsensor               & 0.26$_{\pm{0.11}}$        & 440.17$_{\pm{142.98}}$    & 0.29$_{\pm{0.15}}$        & 516.06$_{\pm{164.20}}$    & 0.50$_{\pm{0.08}}$        & 1243.43$_{\pm{162.25}}$   \\
\midrule
Ours                    & \textbf{0.72$_{\pm{0.07}}$} & \textbf{893.28$_{\pm{65.94}}$} & \textbf{0.69$_{\pm{0.12}}$} & \textbf{953.99$_{\pm{66.65}}$} & \textbf{0.82$_{\pm{0.06}}$} & \textbf{2001.71$_{\pm{66.57}}$} \\
\bottomrule[1pt]  
\end{tabular}%
}}
\vspace{-7mm}
\label{table:results-sim}
\end{table*}

\begin{figure*}[htbp]
    \centering
    \includegraphics[width=0.92\linewidth]{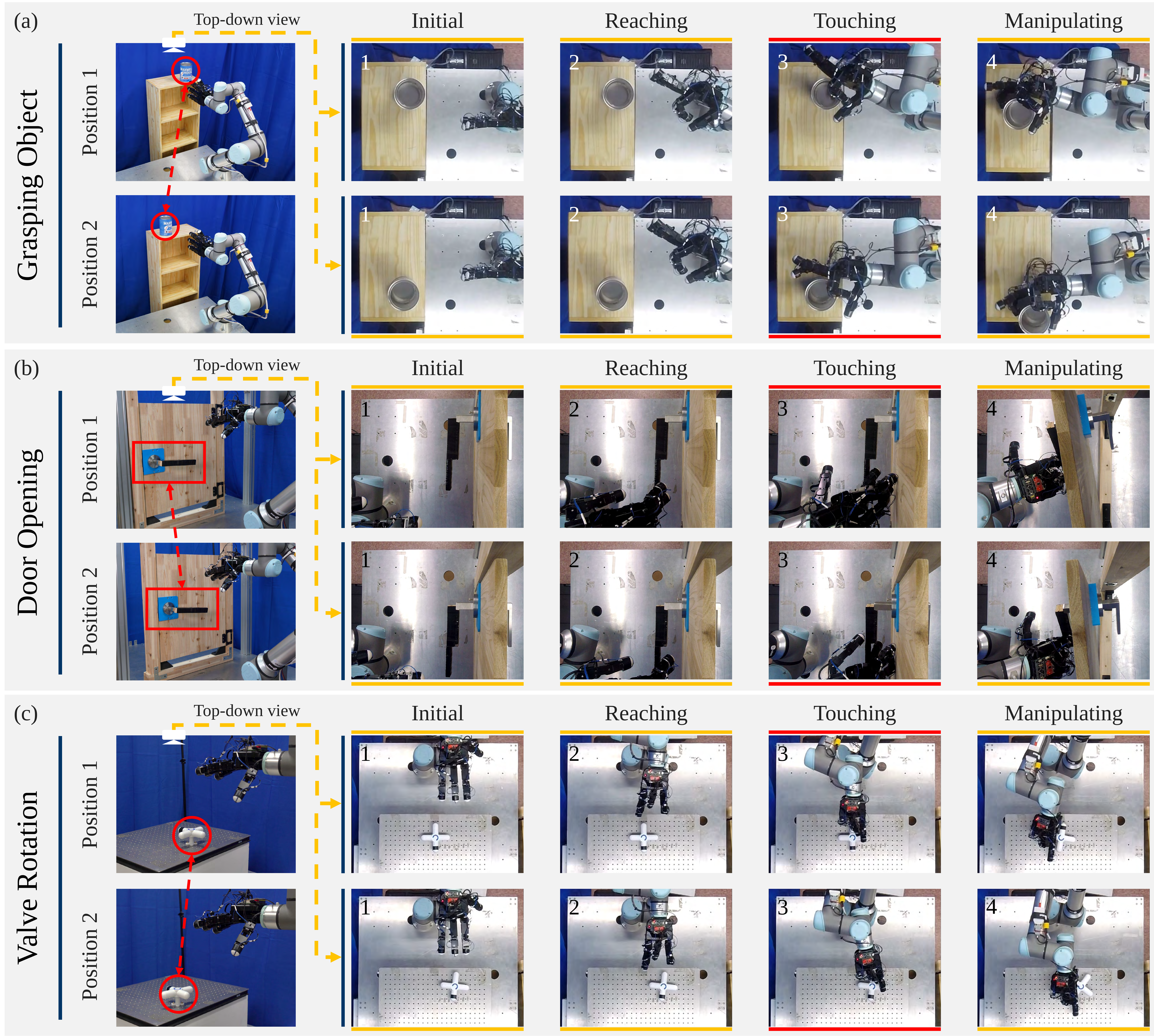}
    \caption{Visualization of the results of performing each task from a top view in the real-world experiment. The red arrow indicates the difference in the position of the target object. Frames illustrating a series of operations demonstrate that the policy can manipulate randomly located objects according to the objectives.
    }
    \label{fig:real_frame}
    \vspace{-5mm}
\end{figure*}

(i) WO-Sensor, which does not have tactile information, learns by leveraging the location information known in the initial training stage (e.g., position of the table in the case of the grasping object task). However, after that, it fails to learn interactions that involve manipulating the target object to succeed in the task. These results show that it is difficult to detect the position of an object during interaction using a robot's proprioception alone, and that tactile sensors play an important role in object detection.

(ii) High-quality tactile sensors enhance the success rates of dexterous manipulation tasks. As the sensitivity of the tactile sensor decreases, more force is required for object contact detection. Phenomena such as objects falling or being thrown have been observed during this process. These findings are consistent with previous results showing that highly sensitive tactile sensors aid in reliable object manipulation.

(iii) ‘WO-PInfo’ achieved lower performance than other methods that can utilize privileged information. This is consistent with the general result that using asymmetric observation improves the learning efficiency of reinforcement learning systems. Privileged information is not accessible to the policy network, however it contributes to the efficiency and stability of learning.

\subsection{Ablation Study I: Analysis of the role of sensors}
Ablation studies were conducted to ascertain the tactile sensor most beneficial for manipulation in the absence of vision.
Two groups were set for comparison. One group, Fingertips, activates only the sensors on the fingertips, while the other group, Palm, activates only the sensors attached to the palm. Both groups have four activated sensors each, and the touch detection threshold was set to 0.01 $\mathrm{N}$, the same as ours. These were trained in the same simulation environment and then compared to our policy. The results are shown in Fig.{~\ref{fig:result_plot}} (b) and Table{~\ref{table:results-sim}}.
The higher success rate of the fingertips group compared to the palm group indicates that fingertip sensors are more advantageous for object manipulation. In particular, this group achieved similar success rates to ours in the Valve Rotating task because the fingertips were primarily used for valve manipulation behavior, minimizing the impact of deactivating other sensors.

\subsection{Ablation Study II: Analysis of the usability of sensor types}
Force/Torque (F/T) sensors are widely used sensors in robot manipulation. For this reason, they are often considered a good alternative to tactile sensors. We conducted ablation studies to ascertain which type of sensor is more crucial for object manipulation without vision. A new group using F/T sensors on the robot wrist was constructed (F/Tsensor). The 3-axis force and torque were measured using the wrist-mounted F/T sensor and utilized for learning. F/Tsensor was also trained in the same simulation environment and compared with our policy, Fingertips, and Palm. The results are shown in Fig.{~\ref{fig:result_plot}} and Table{~\ref{table:results-sim}}.
We observed that F/Tsensor achieved lower success rates compared to methods using touch sensors. This suggests that touch sensors attached to the robot hand are more essential than F/T sensors for navigating and manipulating objects without vision.

\begin{table}
\vspace{2mm}
\caption{
Performance on each task in real-world. The results are averaged over 3 policies trained on 3 seeds.}
\vspace{-1mm}
\resizebox{\columnwidth}{!}{%
% \colorbox{yellow}{
\begin{tabular}{lccc}
\toprule[1.2pt]
Method    & Grasping Object             & Door Opening                & Valve Rotation \\
\midrule
LQ-Sensor & 0.27$_{\pm{0.12}}$          & 0.35$_{\pm{0.13}}$          & 0.32$_{\pm{0.18}}$          \\
DA-Sensor & 0.09$_{\pm{0.04}}$          & 0.11$_{\pm{0.03}}$          & 0.12$_{\pm{0.05}}$          \\
Ours      & \begin{tabular}[c]{@{}c@{}}\textbf{Seen: 0.64$_{\pm{0.17}}$}\\ \textbf{Unseen : 0.47$_{\pm{0.24}}$}\end{tabular} & \textbf{0.60$_{\pm{0.17}}$} & \textbf{0.67$_{\pm{0.17}}$} \\

\bottomrule[1.2pt]
\end{tabular}%
}
\label{table:results-real}
\vspace{-7mm}
\end{table}

\subsection{Real-world evaluation with different sensing capabilities}
We transfer the trained policies to a real robot without fine-tuning to test whether tactile sensors continue to provide benefits for object manipulation without vision.
The results are shown in Table{~\ref{table:results-real}}.
We further analyze the tactile policy, as shown in Fig.{~\ref{fig:real_frame}}, which illustrates the object manipulation process for each task. In particular, the analysis focuses on the transferability of the ability to manipulate randomly located objects.

In all tasks, the initial reaching process was observed to have similar trajectories. This means that in the initial stages, when information is lacking, approaches are based on what is known. However, in subsequent manipulation behaviors, different performance was observed depending on the extent to which touch was used.

Due to the low sensitivity of the LQ-Sensor, more contact force was required for manipulation. As a result, excessive manipulation, such as dropping objects, has been observed. These observations indicate that sensitive touch is important for the manipulation of objects. The lowest success rate of the DA-Sensor indicates a significant contribution of the touch sensor to manipulation. As a result, inital process similar to our policy was performed, but the lack of a touch sensor prevented the object from being found, making subsequent manipulation tasks difficult.

In contrast, our approach leveraging touch provides benefits for object manipulation. In each task, manipulation attempts were observed based on object contact. For example, it has been observed that the timing of attempted manipulation of an object varies depending on when contact occurs. 
This suggests that the robot can first move to an area where an object may exist based on prior information, and then manipulate the object without vision based on touch. Additionally, these observations highlight that sensitive touch can be useful for blind manipulation of objects.

Prior information serves as the starting point for the robot to recognize the existence of an object and to perform the task. From this perspective, prior information can influence the initial stage of object manipulation. However, since the robot cannot determine the exact location of the object with prior information alone, it must learn to manipulate the object by compensating for the loss of information with tactile sense. In addition, the properties of objects encountered by the robot during learning can also be included in the prior information. We observed that, among the three unknown objects used in the Grasping object, the tumbler had the lowest success rate. This is due to the heavy weight and slippery surface of the tumbler. Therefore, it suggests that properties not encountered during learning can affect the task success.

The results demonstrated that policies trained in simulation can be successfully transferred to real-world environments. Even in a real environment, the policy is capable of manipulating randomly located target objects according to the task. This indicates that blind manipulation using tactile information can be effectively applied in real-world scenarios, even in the absence of visual information.

\section{Conclusion}
In this paper, a tactile-based approach has been proposed to enable robots to perform manipulation tasks without vision. Tasks that mimic daily life were defined, and a reinforcement learning framework was presented to learn complex multi-degree-of-freedom movements of the robot arm and hand for each task. 
The proposed system demonstrated that tactile information can be used to enable robots to perform blind manipulation. The results show that the success rate increases as the tactile sensor becomes more sensitive and is distributed over a wider area.
This suggests that objects can be manipulated without relying on visual information, and proves that tactile information is useful for object manipulation in environments where visual information is absent. Future research directions are focused on applying sensors that can provide various tactile information, such as 3-axis force sensors, and expanding the system to study generalizability.

\bibliographystyle{IEEEtran}
\bibliography{IEEEabrv, reference}

\end{document}